\documentclass{article}
\usepackage{spconf,amsmath,graphicx}
\usepackage{epsfig}
\usepackage{graphicx}
\usepackage{amsmath}
\usepackage{amssymb}
\usepackage{algorithm}
\usepackage{algpseudocode}
\usepackage{mathrsfs}

\title{Structured Analysis Dictionary Learning for Image Classification}
\name{Wen Tang, Ashkan Panahi, Hamid Krim, Liyi Dai$^\dagger$ \sthanks{Thanks to US-ARO agreement W911NF-16-2-0005. This is the final version accepted by ICASSP 2018.
	}}
\address{Department of Electrical and Computer Engineering\\
	North Carolina State University, Raleigh, NC, USA\\
	$^\dagger$Army Research Office, RTP, Raleigh, NC, USA\\
	{\tt\small \{wtang6, apanahi, ahk\}@ncsu.edu, liyi.dai@us.army.mil}}
\begin{document}
%
\maketitle
\begin{abstract}
	We propose a computationally efficient and high-performance classification algorithm  by incorporating class structural information in analysis dictionary learning. To achieve more consistent classification, we associate a class characteristic structure of independent subspaces and impose it on the classification error constrained analysis dictionary learning. Experiments demonstrate that our method achieves a comparable or better performance than the state-of-the-art algorithms in a variety of visual classification tasks. In addition, our method greatly reduces the training and testing computational complexity.
\end{abstract}
\section{Introduction}
\label{sec:intro}

Sparse representation has been successfully applied in various image processing and computer vision problems, such as image denoising, and image restoration. Dictionary learning is one way of obtaining sparse representations for signals with unknown precise model. The resulting sparse representation as a linear combination of atoms varies according to the type of dictionary learning techniques: Synthesis Dictionary Learning(SDL) and Analysis Dictionary Learning(ADL). \par
In contrast to SDL, which assumes that the interesting signal can be recovered by a dictionary with corresponding sparse coefficients, ADL is based on applying the dictionary to the data to yield sparse coefficients.


Due to the success of dictionary learning in image restoration problems, task-driven dictionary learning methods are of great interest in many inference problems, such as image classification. 
There are broadly two strategies to address the task-driven dictionary learning method. The first strategy is to learn multiple class-specific sub-dictionaries to make the dictionary more structured, and to increase overall discrimination between different classes \cite{src,ramirez2010classification,yang2011fisher,wang2013max}. To be structured, the atoms in the dictionary are made to learn their own class labels. A class label for a new image can then be decided by comparing reconstruction error from different classes. 
Another strategy is to learn a shared dictionary for all classes and jointly learn a universal classifier to enforce more discriminative sparse representations \cite{mairal2009supervised,lcksvd}. 

All of the above mentioned techniques have been developed and implemented in the SDL framework, while ADL has increasingly received attention\cite{nam2013cosparse}. To the best of our knowledge, none of the standard ADL algorithm such as the analysis K-SVD\cite{Rubi13} or the Sparse Null Space(SNS) pursuit \cite{bian2016sparsity} has addressed the task driven ADL problem. Shekhar \emph{et al.} \cite{shekhar2014analysis} have adopted ADL together with SVM to digits and face recognition, and demonstrated that ADL is more stable under noise and occlusion with a competitive performance with SDL. Guo \emph{et al.} \cite{guo2016discriminative} integrated local topological structures and discriminative sparse labels into the ADL to yield a $k$ Nearest Neighbor method to classify images.

Inspired by these past efforts and efficient coding of ADL, we propose an integration of structured subspace regularization and supervised learning into an ADL model to obtain a more structured discriminative and efficient approach to image classification.
It has been shown, for example in the context of sparse subspace clustering \cite{elhamifar2013sparse}, that the sparse representations of the data within a class share a low dimensional subspace. A structuring block diagonal matrix therefore is introduced to achieve these localized subspaces of the sparse codes. This yields more coherence for within-class sparse representations and more disparity for between-class representations. To induce additional robustness in the sought sparse representation, a one-against-all regression-based classifier is jointly learned, with a resulting optimization functional which we solve by a linearized alternating direction method (ADM)\cite{lin2011linearized}. This approach is computationally more efficient than analysis K-SVD\cite{Rubi13} and SNS pursuit \cite{bian2016sparsity}.
Moreover, a great advantage of our algorithm is its extremely short on-line encoding and classification time. 
Our experiments demonstrate that our method achieves a better overall performances than the synthesis dictionary approach.

The balance of this paper is organized as follows: In Section \ref{sec:SSADL}, we state and formulate the problem. We discuss the resulting solution to the optimization problem in Section \ref{sec:solve}. 
The experimental validation and results are comprehensively presented in Section \ref{sec:experiments}. We finally provide some concluding comments in Section \ref{Conclusion1}.

\section{Structured Analysis Dictionary Learning}
\label{sec:SSADL}
\noindent\textbf{\textit{Notation:}} Uppercase and lowercase letters respectively denote matrix and vectors. The transpose and inverse of matrix are represented as the superscripts $T$ and $-1$, such as $A^T$ and $A^{-1}$. $(a_i)_j$ represents the $j$th element in the $i$th column of matrix $A$.

\subsection{ADL Formulation}
Given a data matrix $X=[x_1,\dots,x_n] \in \mathbb{R}^{m \times n}$, the originally formulated ADL\cite{Rubi13} problem seeks a representation frame $\Omega$ with a sparse coefficient set $U$.
\begin{equation}\label{equ:adl}
\begin{split}
\arg\min_{\Omega,U} & ~ \frac{1}{2}\|U-\Omega X\|_2^2+\lambda_1 \|U\|_1\\
s.t. &~ \Omega \in \mathbb{R}^{r\times m} \subset \mathcal{W},
\end{split}
\end{equation}
where $U \in \mathbb{R}^{r \times n}$ and $\mathcal{W}$ is a non-trivial solution set.

\subsection{Mitigating Inter-Class Feature Interference}
The basic idea in our algorithm is to employ the representation $U$ to obtain a classifier. To reduce the impact of inter-class common atoms on the discriminative power of ADL, we propose two additional constraints on $U$ by way of: (1) A structural map of $U$ to minimize interference of inter-class common features. (2) A classification error performance minimization.
\\
\noindent\textbf{\textit{(1) Structural Mapping of U}:} This constraint is particularly enforced by imposing that each class belongs to a subspace defined by a span of the associated coefficients. This improves the consistency of the analysis representations within a class and enhances the divergence between different classes. A block-diagonal matrix $H \in \mathbb{R}^{s \times n}$ as shown below is hence introduced in the training phase,
\[H=
\bordermatrix{~& h^1_1 & h^1_2 & h^1_3 & h^2_4 & h^2_5 \cr
	~& 1 & 1 & 1 & 0 & 0 \cr
	~& 1 & 1 & 1 & 0 & 0  \cr
	~& 1 & 1 & 1 & 0 & 0  \cr
	~& 0 & 0 & 0 & 1 & 1 \cr
	~& 0 & 0 & 0 & 1 & 1 \cr
	}
,\]
where $s\geq n$ is the length of structured representation. Each diagonal block represents a class and each column $h_i^j$ is a structured representation for the corresponding data point $i$ in the $j$th class. 
This constraint may also be deviated by an error term, to be jointly minimized with the ADL functional,
\begin{equation}
\label{equ:ssp}
H=QU+\varepsilon_1,
\end{equation}
where $Q \in \mathbb{R}^{s \times r} $ is matrix to be learned with $\Omega$ and $U$, $\varepsilon_1$ is the tolerance. 
\\
\noindent\textbf{\textit{(2) Minimal Classification Error}:}
The second constraint is a classification error as a feedback term to the learning process of $\Omega$ and $U$. A regression-based classifier $W \in \mathbb{R}^{c \times s}$ is applied to the structured representations $QU$ in this term. We write it as
\begin{equation}
\label{equ:lc}	
L=W(QU)+\varepsilon_2,
\end{equation}
where $\varepsilon_2$ is also the tolerance, and the label matrix $L \in \mathbb{R}^{c \times n}$, with $c$ denoting for the number of classes. If image $j$ \text{ belongs to class } $i,$ $L_{ij}=1$; otherwise, $L_{ij}=0$.

\par
\subsection{Structured ADL Formulation}
To ensure that the structure for each image class is preserved together with minimal interference between different classes, the minimization of tolerance errors is also required. Then, using Eqs.(\ref{equ:ssp}), (\ref{equ:lc}) and the minimization of tolerance errors together, the resulting algorithm formulation for our structured ADL is written as 
\begin{equation}\label{equ:ssadl}
\begin{split}
\arg \min_{\substack{\Omega, U, Q, \\ W, \varepsilon_1, \varepsilon_2}} & \frac{1}{2}\|U - \Omega X\|_F^2+ \lambda_1  \|U\|_1\\
&+\frac{\rho_1}{2}\|\varepsilon_1\|^2_2+\frac{\rho_2}{2}\|\varepsilon_2\|^2_2\\
\emph{s.t.} ~&H=QU+\varepsilon_1,\\
& L=W(QU)+\varepsilon_2,\\
&\|\omega_i^T\|_2^2=1; \forall i=1,\dots,r,
\end{split}
\end{equation}
where $\omega_i^T$ is the row of $\Omega$, $\rho_1$ and $\rho_2$ are the penalty coefficients. Recall $H$ is the structured representation, $Q$ is the structuring transformation, $L$ is the classifier label, and $W$ is the linear classifier, and 
$\lambda_1$ is the tuning parameters.\par



\section{Algorithmic Solution}\label{sec:solve}
The objective function in Eq.(\ref{equ:ssadl}), on account of its non-convexity, is transformed to an augmented Lagrange formulation with dual variables $Y^{(1)}$, $Y^{(2)}$ and $\mu$. After straight forward calculations that lead to eliminations of $\varepsilon_1$ and $\varepsilon_2$, we obtain the following expression for this function:
\begin{equation}\label{equ:Lssadl}
\begin{split}
&L(\Omega, U, Q, W, Y^{(1)}, Y^{(2)}, \mu)=\frac{1}{2}\|U - \Omega X\|_F^2 +\lambda_1 \|U\|_1  \\
&+ \lambda_2 <Y^{(1)}, H-QU>+\lambda_3 <Y^{(2)},L-W(QU)>\\
&+\frac{\mu}{2}\|H-QU\|_2^2+\frac{\mu}{2}\|L-W(QU)\|_2^2,
\end{split}
\end{equation}
where $\lambda_1$, $\lambda_2$, $\lambda_3>0$ are the new tuning parameters. 
Then, to minimize the objective functional in Eq.(\ref{equ:Lssadl}), we first randomly initialize the analysis dictionary $\Omega$ and two linear transformations $Q$ and $W$. The sparse representation $U$ is initialized by $U=\textbf{0}$, the zero matrix. $\eta_{Q}$, $\eta_{WQ}$, and $\eta_{WU} >0$ are the parameters for the learning rate. Then, we alternately update different variables when fixing the others, which is summarized in Algorithm 1.\par

\renewcommand{\algorithmicrequire}{\textbf{Input:}}
\renewcommand{\algorithmicensure}{\textbf{Output:}}

\begin{algorithm} \label{alg:learning}  
	\caption{Structured Analysis Dictionary Learning}  
	\begin{algorithmic}[1] 
		\State Initialize $\Omega$, $Q$, and $W$ as random matrices, and initialize $U$ as a zero matrix; $T$ is maximum iteration;
		\While {not converged \textbf{and} $k < T$} 
		\State $k=k+1;$
		\State \resizebox{0.9\columnwidth}{!}{$U_{k+1}=\tau_{\frac{\lambda_1}{\mu_k(\eta_Q+\eta_{WQ})}} \left(U_k-\frac{\bigtriangledown_UL(\Omega_k,U_k,Q_k,W_k,Y^{(1)}_k,Y^{(2)}_k)}{\mu_k (\eta_Q+\eta_{WQ})}\right);$}
		\State $Q_{k+1}=Q_k-\frac{\bigtriangledown_QL(\Omega_k,U_{k+1},Q_k,W_k,Y^{(1)}_k,Y^{(2)}_k) }{\mu_k(\eta_Q+\eta_{WU})},$
		\State $W_{k+1}=W_k-\frac{\bigtriangledown_WL(\Omega_k,U_{k+1},Q_{k+1},W_k,Y^{(1)}_k,Y^{(2)}_k) }{\mu_k\eta_{QU}};$
		\State $\Omega_{k+1}=U_{k+1}X^T(XX^T+\lambda_4 I)^{-1};$
		\State Normalize $\Omega_{k+1}$ by $\omega_i^T=\frac{\omega_i^T}{\|\omega_i^T\|_2}, \forall i;$
		\State $Y_{k+1}^{(1)}=Y_{k}^{(1)}+\mu_k(H-Q_{k+1}U_{k+1});$
		\State $Y_{k+1}^{(2)}=Y_{k}^{(2)}+\mu_k(L-W_{k+1}Q_{k+1}U_{k+1});$
		\State $\mu_{k+1}=\min\{\rho\mu_k,\mu_{max}\};$ $\%\rho$ is the learning rate
		\EndWhile
	\end{algorithmic}  
\end{algorithm}  
\section{Experiments and Results}\label{sec:experiments}
We evaluate our proposed SADL method on four popular visual classification datasets which have been widely used in previous works and with known performance benchmarks. They include Extended YaleB\cite{yaleB} face dataset, AR\cite{AR} face dataset, Caltech101\cite{caltech101} object categorization dataset and Scene15\cite{scene15} scene image dataset. The features of these 4 datasets are extracted by the same settings in \cite{lcksvd}.\par

In our experiments, we provide a comparative evaluation of three state-of-the-art techniques and our proposed technique, including classification accuracy and training and testing times. 
The testing time is defined as the average processing time to classify a single image. 
For a fair comparison, we measure the performances of all algorithms by using the same dictionary size on each dataset and experiment over 10 realizations to obtain an average performance. In relation to competitive methods, ADL+SVM \cite{shekhar2014analysis} is a baseline. SRC \cite{src} is the classical Sparse Representation based Classification. LC-KSVD \cite{lcksvd} is a SDL approach that jointly learns a discriminative dictionary and a universal classifier. In our tables, the accuracy in the parentheses with the citation is the one that was reported in the original paper. The difference of the accuracy of our implementing and the original one might be caused by the different segmentations of the training and testing samples.

\subsection{Face Recognition}
\begin{figure}[htb]
	\centering
	\includegraphics[width=0.45\textwidth]{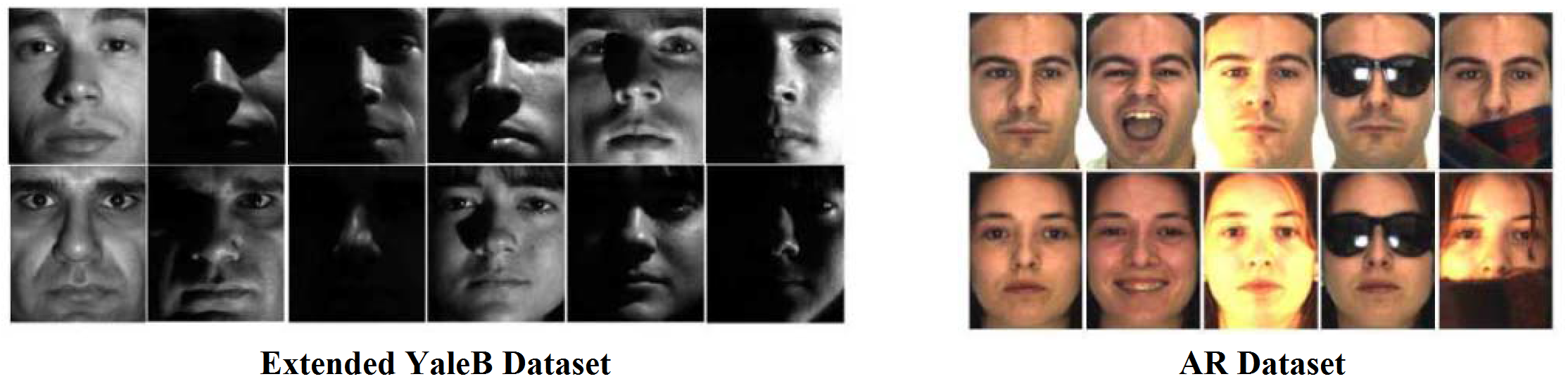}
	\caption{Examples of Face Dataset: The left figure is Extended YaleB Dataset, and the right one is AR Dataset.}
	\label{fig:YaleB}
\end{figure}
\noindent\textbf{\textit{Extended YaleB}:} This face dataset contains in total 2414 frontal face images of 38 persons under various illumination and expression conditions, as illustrated in Fig.\ref{fig:YaleB}. 
Each Extended YaleB face image has a $504$-dimensional feature vector. We randomly choose half of the images for training, and the rest for testing. The dictionary size is set to 570 atoms, $\lambda_1=0.001$, $\lambda_2=9$, $\lambda_3=3$, $\lambda_4=0.5$ and $T=780$.

\begin{table}[htb]
	\centering
	\caption{Classification Results on Extended YaleB Dataset}
	\resizebox{\columnwidth}{!}{%
	\begin{tabular}{llll}
		\hline
		Methods & Accuracy (\%) & Training (s) & Testing (s)\\
		\hline
		ADL+SVM\cite{shekhar2014analysis} & 82.91\% & 91.78 & 1.13$\times 10^{-3}$\\
		SRC\cite{src} & 80.5\% & No Need & 3.74$\times 10^{-1}$\\
		LC-KSVD\cite{lcksvd} & 94.56\% (\textbf{95}\% \cite{lcksvd}) & 234.67 & 1.63$\times 10^{-2}$\\
		SADL & 94.91\% &\textbf{51.29} & \textbf{2.72$\times 10^{-6}$}\\
		\hline
	\end{tabular}%
	}
	\label{tab:yaleb}%
\end{table}%

The classification results, training and testing times are summarized in Table \ref{tab:yaleb}. Our proposed SADL method achieves the highest classification accuracy in the test, but tinily lower than the reported accuracy of LC-KSVD. However, it is still substantially more efficient than the others in terms of numerical complexity and classification . \par

\begin{figure}[htb]
	\centering
	\includegraphics[width=0.40\textwidth]{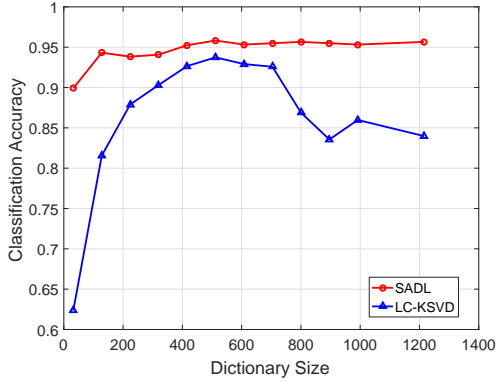}
	\caption{Classification Accuracy}
	\label{fig:dictsize}
\end{figure}

For a more thorough evaluation, we compare SADL with LC-KSVD for different dictionary sizes, and display the classification accuracy in Fig.\ref{fig:dictsize}. We ran our experiments for dictionary sizes by 32, 128, 224, 320, 416, 512, 608, 704, 800, 896, 992, and 1216 (all training size). SADL exhibits a more stable performance than that of LC-KSVD. In particular, the accuracy of LC-KSVD significantly decreases, when the dictionary size approaches the all training sample size. In addition, our method apparently has a much higher classification accuracy than LC-KSVD, when the dictionary size is small. The significant decrease in accuracy may be caused by the trivial solution of dictionary in SDL. 

\noindent\textbf{\textit{AR}:} The AR face dateset has 2600 color images of 50 females and 50 males with more facial variations than the Extended YaleB database, such as different illumination conditions, expressions and facial disguises, as shown in Fig. \ref{fig:YaleB}. Each person has about 26 images of size $165 \times 120$. 
The AR Face feature dimension is 540. 20 images of each person are randomly selected as a training set and the other 6 images for testing. The dictionary size of the AR dataset is set to 500 atoms, $\lambda_1=0.001$, $\lambda_2=8$, $\lambda_3=10$, $\lambda_4=0.5$ and $T=1040$.

The classification performances are summarized in Table \ref{tab:ar}. Our proposed SADL achieves higher classification accuracy than others. Our method is about 10000 times faster than SRC and LC-KSVD for the testing phase.

\begin{table}[htb]
	\centering
	\caption{Classification Results on AR Dataset}
		\resizebox{\columnwidth}{!}{%
	\begin{tabular}{llll}
		\hline
		Methods & Accuracy (\%) & Training (s) & Testing (s)\\
		\hline
		ADL+SVM\cite{shekhar2014analysis} & 90.40\% & 218.54 & 9.10$\times 10^{-3}$\\
		SRC\cite{src} & 66.50\% & No Need & 5.25$\times 10^{-2}$\\
		LC-KSVD\cite{lcksvd} & 87.78\% (93.7\%\cite{lcksvd}) & 244.52 & 1.42$\times 10^{-2}$\\
		SADL & \textbf{95.08\%} & \textbf{89.13} & \textbf{3.67$\times 10^{-6}$}\\
		\hline
	\end{tabular}%
	}
	\label{tab:ar}%
\end{table}%

\subsection{Object Recognition}
\begin{figure}[htb]
	\centering
	\includegraphics[width=0.31\textwidth]{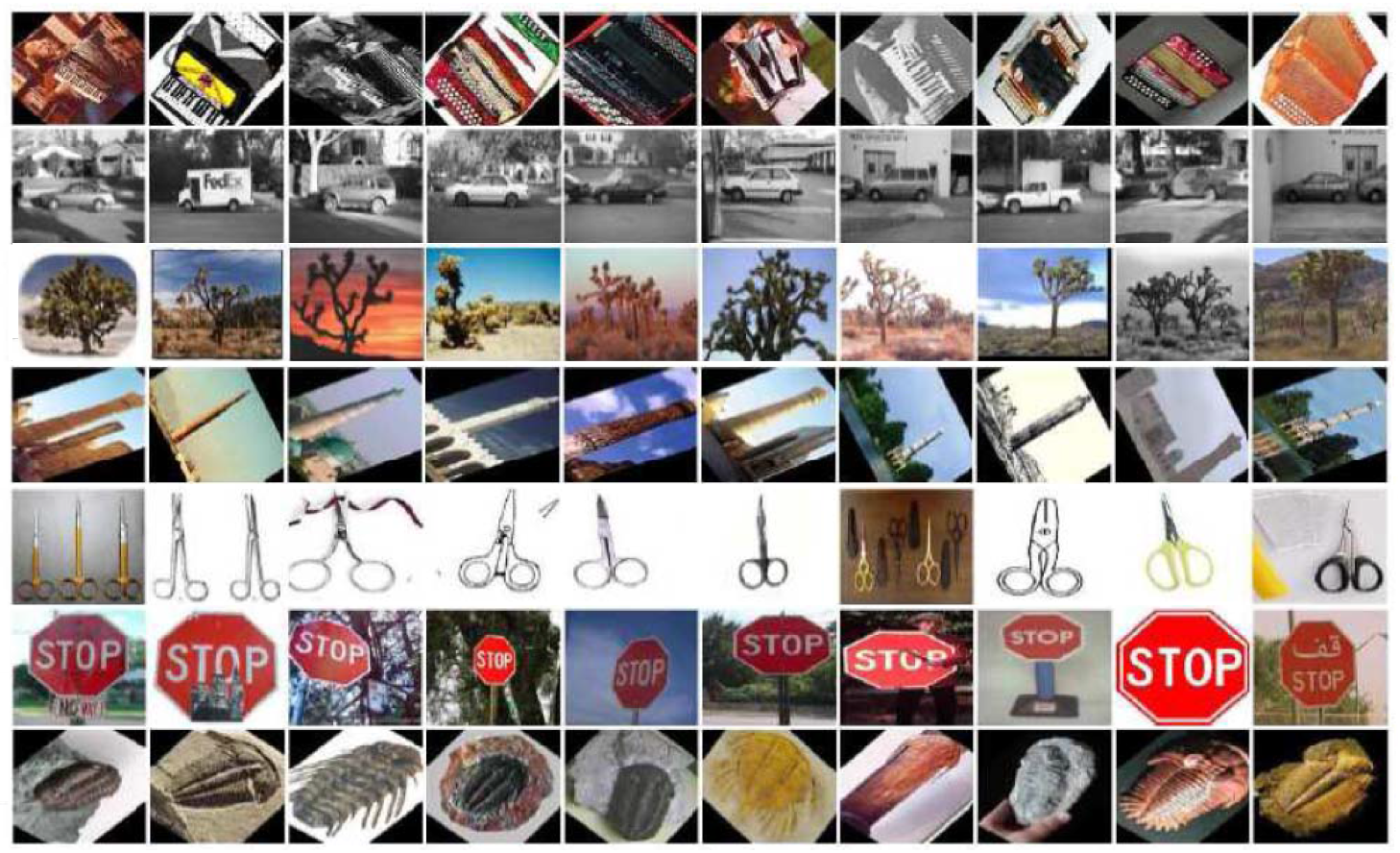}
	\caption{Caltech101 Dataset Examples}
	\label{fig:caltech101}
\end{figure}

The Caltech101 dataset has 101 different categories of different objects and 1 non-object category. Most categories have around 50 images. Fig.\ref{fig:caltech101} gives some examples from the Caltech101 dataset. The standard bag-of words+spatial pyramid matching (SPM) framework \cite{scene15} is used to calculate the SPM features. PCA is then adopted to reduce the dimension of a SPM feature to 3000. 
The dictionary size is set to 510, $\lambda_1=0.001$, $\lambda_2=10$, $\lambda_3=3$, $\lambda_4=4.6$ and $T=990$.

\begin{table}[htb]
	\centering
	\caption{Classification Results on Caltech101 Dataset}
		\resizebox{\columnwidth}{!}{%
	\begin{tabular}{llll}
		\hline
		Methods & Accuracy (\%) & Training (s) & Testing (s)\\
		\hline
		ADL+SVM\cite{shekhar2014analysis} & 54.93\% &\textbf{447.80}  & 7.75$\times 10^{-3}$\\
		SRC\cite{src} & 67.70\% & No Need & 4.34$\times 10^{-1}$\\
		LC-KSVD\cite{lcksvd} & 71.79\% & 487.61 & 1.35$\times 10^{-2}$\\
		SADL & \textbf{72.36\%} & 773.66 & \textbf{8.10$\times 10^{-6}$}\\
		\hline
	\end{tabular}%
	}
	\label{tab:caltech101}%
\end{table}%

We evaluate all methods with a dictionary size of 510. The classification performances are summarized in Table \ref{tab:caltech101}. Our proposed SADL still achieves the highest performance of the lot. SADL has again a short testing time, which is around 10000 times faster than LC-KSVD.

\subsection{Scene Classification}
\begin{figure}[htb]
	\centering
	\includegraphics[width=0.35\textwidth]{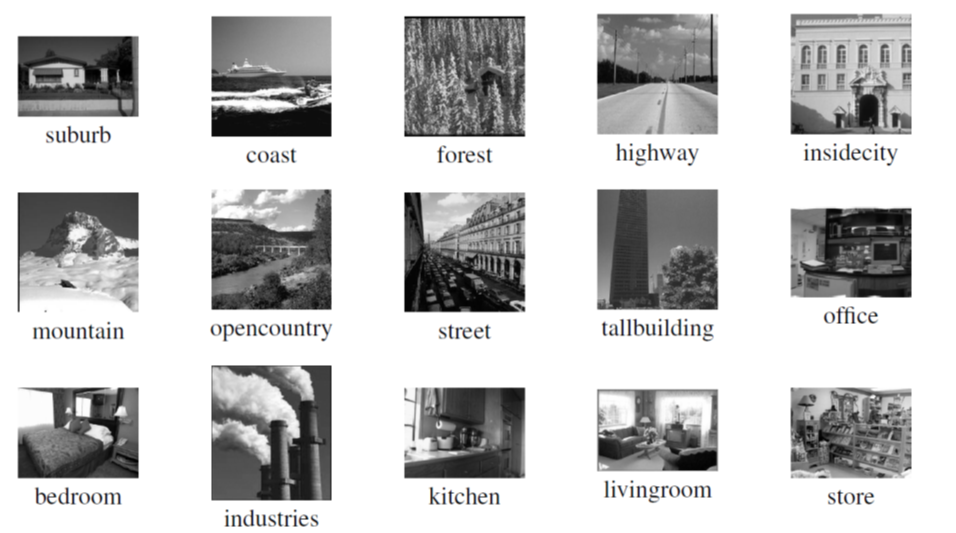}
	\caption{Scene15 Dataset Examples}
	\label{fig:scene15}
\end{figure}
Scene15 dataset contains a total of 15 categories of different scenes, and each category has around 200 images. The examples are listed in Fig.\ref{fig:scene15}. Proceeding as for the Caltech 101 dataset, we compute the SPM features for scene images. Each scene image is transformed to a 3000 dimensional feature by PCA. We randomly pick 100 images per class as training data, and use the rest of images as testing data. The settings and steps follow \cite{lcksvd}. The dictionary size is set to 450, $\lambda_1=0.001$, $\lambda_2=10$, $\lambda_3=4$, $\lambda_4=0.001$ and $T=220$.

\begin{table}[htb]
	\centering
	\caption{Classification Results on Scene15 Dataset}
		\resizebox{\columnwidth}{!}{%
	\begin{tabular}{llll}
		\hline
		Methods & Accuracy (\%) & Training (s) & Testing (s)\\
		\hline
		ADL+SVM\cite{shekhar2014analysis} & 49.35\% & \textbf{110.47} & 1.14$\times 10^{-4}$\\
		SRC\cite{src} & 91.80\% & No Need & 4.06$\times 10^{-1}$\\
		LC-KSVD\cite{lcksvd} & \textbf{98.83\%} (92.9\%\cite{lcksvd}) & 270.93 & 1.26$\times 10^{-2}$\\
		SADL & 98.16\% & 121.02 &\textbf{9.23$\times 10^{-6}$}\\
		\hline
	\end{tabular}%
	}
	\label{tab:scene15}%
\end{table}%

The classification performances are summarized in Table \ref{tab:scene15}. Our performance is slightly lower than LC-KSVD, but is still higher than SRC, ADL+SVM and the LC-KSVD reported accuracy. However, the testing phase is superior to the others. Note that, the testing time is 10 thousand times faster than LC-KSVD. 



\section{Conclusion}\label{Conclusion1}
We proposed an image classification method referred to as structured analysis dictionary learning (SADL). To obtain SADL, we constrained a structured subspace(cluster) model in the enhanced ADL method, where each class was represented by a structured subspace. The enhancement of ADL was realized by constraining the learning by a classification fidelity term on the sparse coefficients. Our formulated optimization problem was efficiently solved by the linearized ADM method, in spite of its non-convexity due to bilinearity. Taking advantage of analysis dictionary, our method achieved a significantly faster testing time.\par



\nocite{*} 

\bibliographystyle{IEEEbib}
\bibliography{egbib}

\end{document}